\documentclass[sigconf]{acmart} 

\usepackage{subfiles} 
\usepackage{multirow}
\usepackage{color}
\usepackage{colortbl}
\definecolor{LightCyan}{rgb}{0.88,1,1}

\AtBeginDocument{%
  \providecommand\BibTeX{{%
    \normalfont B\kern-0.5em{\scshape i\kern-0.25em b}\kern-0.8em\TeX}}}

\copyrightyear{2022}
\acmYear{2022}
\setcopyright{acmcopyright}
\acmConference[MM '22]{Proceedings of the 30th ACM International Conference on Multimedia}{October 10--14, 2022}{Lisboa, Portugal}
\acmBooktitle{Proceedings of the 30th ACM International Conference on Multimedia (MM '22), October 10--14, 2022, Lisboa, Portugal}
\acmPrice{15.00}
\acmDOI{10.1145/3503161.3548123}
\acmISBN{978-1-4503-9203-7/22/10}

\begin{document}
\fancyhead{}
\newcommand{\TODO}[1]{{\color{red}{[TODO: #1]}}}
\newcommand{\zw}[1]{{\color[rgb]{0.7,0.3,0.7}{[ZW: #1]}}}
\newcommand{\jl}[1]{{\color[rgb]{0.7,0.7,0.3}{[JL: #1]}}}
\newcommand{\phil}[1]{{\color[rgb]{0.3,0.7,0.3}{#1}}}
\newcommand{\para}[1]{\vspace{.05in}\noindent\textbf{#1}}
\def\ie{\emph{i.e.}}
\def\eg{\emph{e.g.}}
\def\etal{{\em et al.}}
\def\etc{{\em etc.}}

\title{Boosting Single-Frame 3D Object Detection by Simulating Multi-Frame Point Clouds}

\author{Wu Zheng}
\email{wuzheng@cse.cuhk.edu.hk}
\affiliation{%
  \institution{CSE \& SHIAE, CUHK}
  \city{Hong Kong}
  \country{China}
}

\author{Li Jiang}
\email{lijiang@mpi-inf.mpg.de}
\affiliation{%
  \institution{Max Planck Institute}
  \city{Saarbrücken}
  \country{Germany}
}

\author{FanBin Lu}
\email{1155141511@link.cuhk.edu.hk}
\affiliation{%
  \institution{CSE, CUHK}
  \city{Hong Kong}
  \country{China}
}

\author{Yangyang Ye}
\email{yeyangyang@zju.edu.cn}
\affiliation{%
  \institution{Zhejiang University}
  \city{Hang Zhou}
  \country{China}
}

\author{Chi-Wing Fu}
\email{cwfu@cse.cuhk.edu.hk}
\affiliation{%
  \institution{CSE \& SHIAE, CUHK}
  \city{Hong Kong}
  \country{China}
}

\renewcommand{\shortauthors}{Li and Dai, et al.}

\begin{abstract}
To boost a detector for single-frame 3D object detection, we present a new approach to train it to simulate features and responses following a detector trained on multi-frame point clouds.
Our approach needs multi-frame point clouds only when training the single-frame detector, and once trained, it can detect objects with only single-frame point clouds as inputs during the inference.
For this purpose, we design a novel Simulated Multi-Frame Single-Stage object Detector (SMF-SSD) framework:
multi-view dense object fusion to densify ground-truth objects to generate a multi-frame point cloud;
self-attention voxel distillation to facilitate one-to-many knowledge transfer from multi- to single-frame voxels;
multi-scale BEV feature distillation to transfer knowledge in low-level spatial and high-level semantic BEV features; and
adaptive response distillation to activate single-frame responses of high confidence and accurate localization.
Experimental results on the Waymo test set show that our SMF-SSD {\em consistently\/} outperforms all state-of-the-art single-frame 3D object detectors {\em for all object classes of difficulty levels 1 and 2 in terms of both mAP and mAPH\/}.

\end{abstract}


\maketitle

\section{Introduction}
\begin{figure}
\centering
\includegraphics[width=8cm]{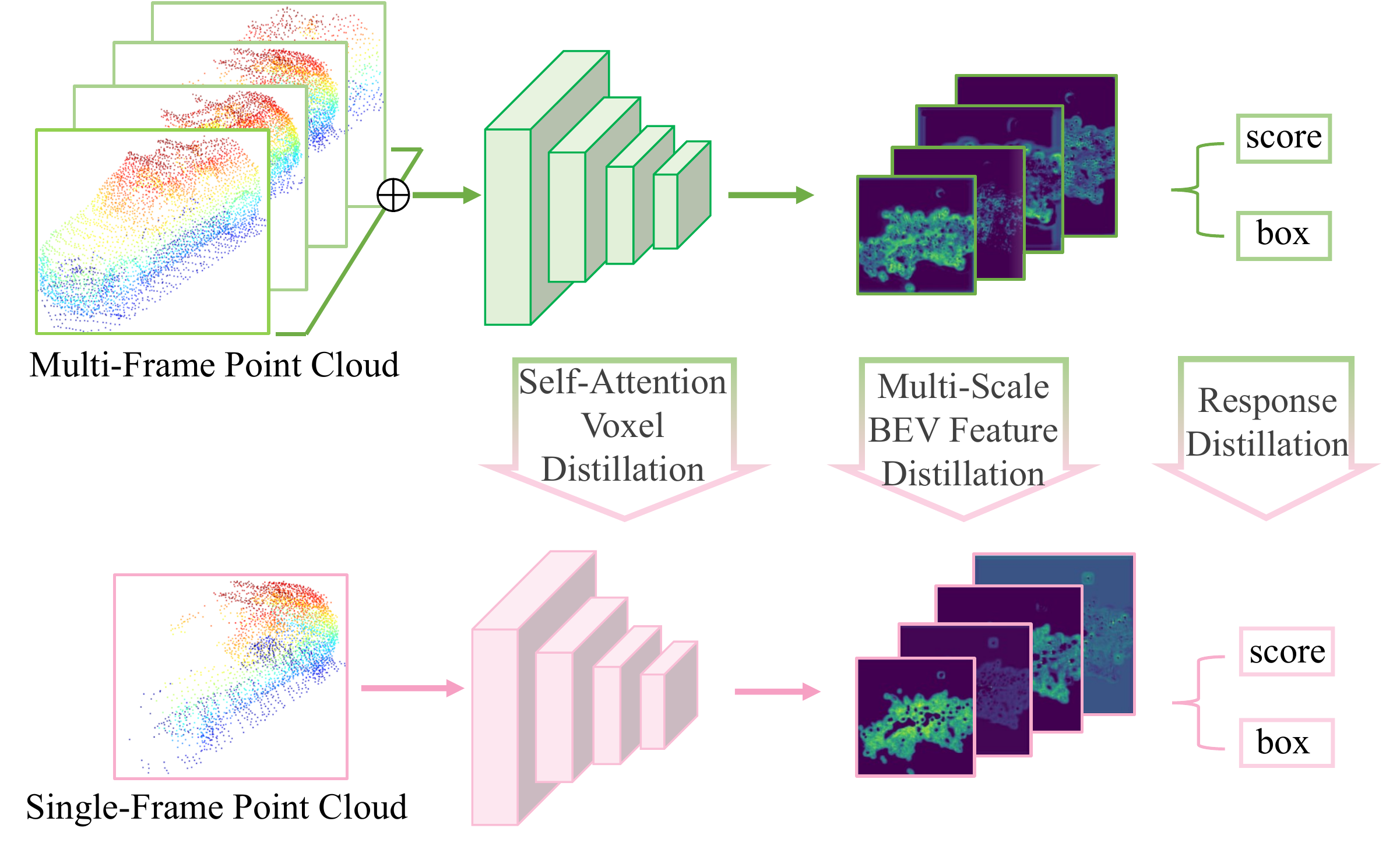}
\caption{
Overview of our Simulated Multi-Frame Single-Stage 3D object Detector (SMF-SSD) framework---we train a single-frame SSD to simulate features and responses following a multi-frame SSD by designing a family of distillation techniques, aiming to boost the single-frame 3D detection precision without sacrificing the inference efficiency.
}
\label{fig:cover}
\vspace*{-1.5mm}
\end{figure}

To support autonomous driving, LiDAR-produced point clouds are widely adopted in state-of-the-art 3D object detectors~\cite{yin2021center,zheng2021se,shi2020pv,he2020structure,yang2019std}, as they are robust to provide precise depth information in varying illumination and weather conditions.
Also, the decreasing cost of LiDAR sensors promotes the popularity of point clouds in self-driving perception.
Yet, due to the self occlusion of objects and the limited number of laser beams produced by LiDAR sensors, object surfaces are unlikely fully covered by the laser rays.

To overcome these shortcomings of 3D sensing with LiDAR, two very recent works~\cite{yang20213d,qi2021offboard} attempt to exploit multi-frame point clouds, meaning that multiple point clouds acquired over a time range are taken together for 3D object detection.
By this means, we may obtain more complete points to cover the object surfaces and improve the detection performance.
3D Auto Labeling~\cite{qi2021offboard} proposes to link the objects localized across multiple point-cloud frames to generate dense object points to further refine the single-frame predictions, whereas
3D-MAN~\cite{yang20213d} proposes to align and aggregate multi-view features extracted from multiple point-cloud frames to enhance the single-frame features.
Though these techniques greatly improve the detection precision, multi-frame detectors inevitably sacrifice the real-time efficiency, since they require 
tremendous computation to process a large volume of point-cloud frames.

For 3D object detection, choosing between single- and multi-frame point clouds is essentially a choice between efficiency and accuracy.
In this work, we attempt to exploit the gap between the two choices, by exploring the possibility of training a single-frame detector with only single-frame point clouds as inputs, yet capable of generating features and responses similar to those from a multi-frame detector.
Our insight behind the idea is that when we see a 3D object from just one single view, we may still infer a more complete picture of the object based on our prior knowledge.
Hence, when we train the single-frame detector, we take a multi-frame detector as its teacher to provide it with more complete 3D knowledge about the single-frame point cloud, such that it can learn to produce reliable features and responses like the teacher.

Our main contribution is a novel Simulated Multi-Frame Single-Stage object Detector (SMF-SSD) framework (see Figure~\ref{fig:cover}) to effectively train a single-frame object detector to produce features and responses similar to those from a multi-frame object detector.
%
%
%
In detail,
%
%
we first design the multi-view dense object fusion to generate light multi-frame point clouds with only dense ground-truth objects, which are aggregated from multi-view object points in sequential time frames.
This mechanism avoids huge computation overhead on the massive points produced from directly concatenating all point-cloud frames.
Second, we formulate the self-attention voxel distillation to build an intermediate domain on single-frame voxels with self attention, and then transfer the knowledge in a one-to-many manner from multi-frame voxels to single-frame ones.
Third, we design a multi-scale region proposal network to abstract 3D voxel features into multi-scale bird's eye view (BEV) features, and then formulate the multi-scale BEV feature distillation on them to separately transfer the knowledge of low-level spatial features and high-level semantics.
Last, we formulate the adaptive response distillation to enhance the optimization of the classification responses with high confidence and the regression responses well aligned with ground-truth bounding boxes, aiming to activate the single-frame predictions of high confidence and accurate localization.

\begin{figure}
\centering
\includegraphics[width=7.8cm]{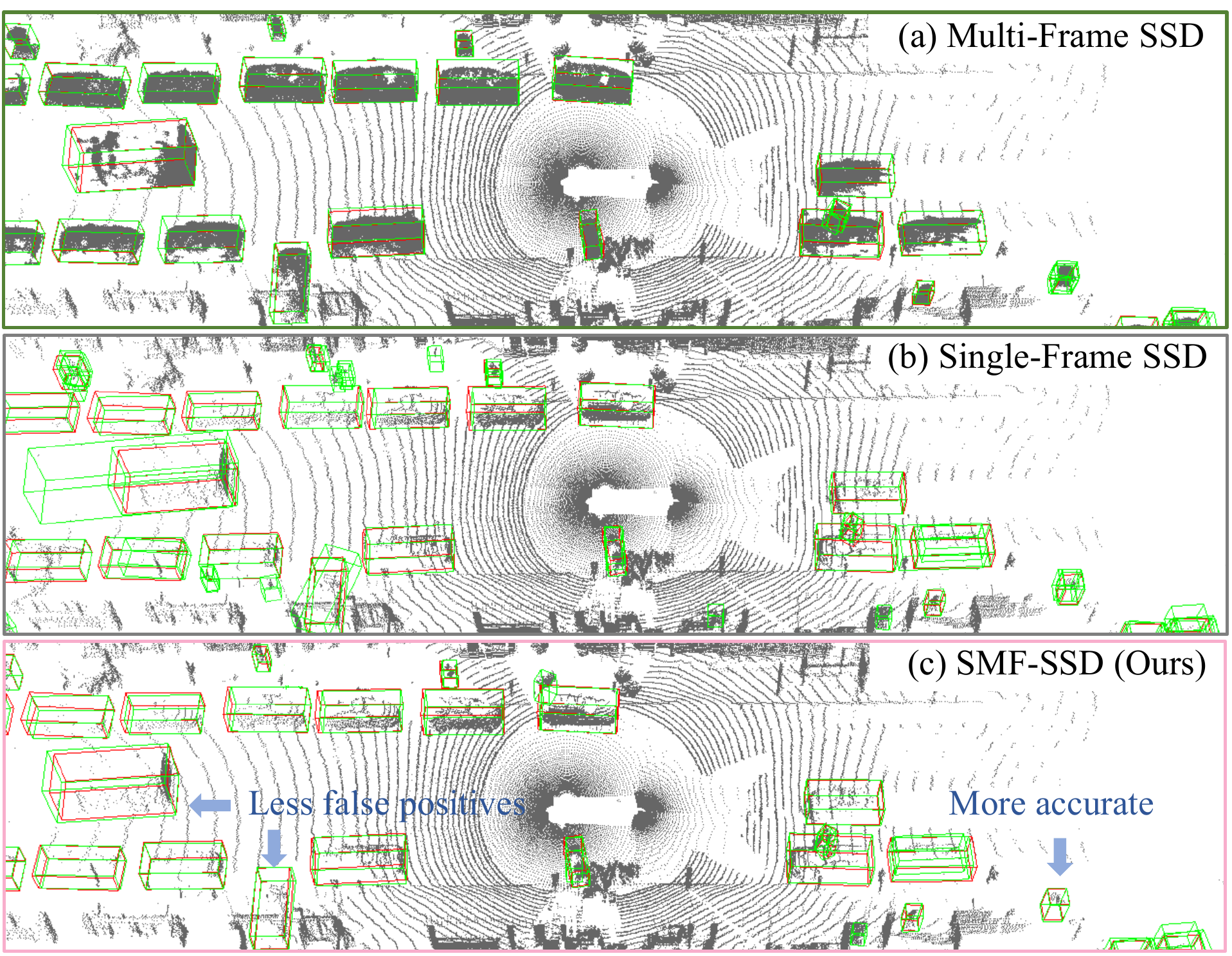}
\vspace*{-1.75mm}
\caption{
Compared to the common single-frame SSD (b), our SMF-SSD (c) predicts bounding boxes (green) that better align with the ground truths (red), while avoiding false predictions, by learning from a multi-frame SSD (a).
}
\label{fig:cover2}
\vspace*{-3mm}
\end{figure}

%
With the above techniques, we can effectively boost the precision of a single-frame detector without slowing down its inference speed; see a detection example with our SMF-SSD in Figure~\ref{fig:cover2}.
The evaluation on the Waymo test set also shows that our SMF-SSD outperforms prior SOTA single-frame detectors on all object classes, and sets a new SOTA performance in terms of the average precision of all classes in both difficulty levels 1 and 2.

\section{Related Work}
\textbf{3D Object Detection.} \
Among the LiDAR-based 3D object detectors,
two-stage detectors~\cite{wang2019frustum,shi2019pointrcnn,yang2019std,shi2020points,shi2020pv,deng2021voxel} focus on improving the region-proposal-aligned features for a better second-stage refinement.
For example, PointRCNN~\cite{shi2019pointrcnn} and STD~\cite{yang2019std} use PointNet~\cite{qi2017pointnet,qi2017pointnet++} to aggregate raw coordinates, semantic and segmentation features of points for region proposals, while Part-$A^2$~\cite{shi2020points}, PV-RCNN~\cite{shi2020pv}, and Voxel R-CNN exploit voxelization or preset grid points to reduce the representation ambiguity of region proposals.
Among the single-stage detectors~\cite{yang20203dssd,he2020structure,zheng2020cia,zheng2021se,yin2021center},
SE-SSD~\cite{zheng2021se} builds a self-ensembling framework with CIA-SSD~\cite{zheng2020cia} to optimize the model with both hard and soft targets.
CenterPoint~\cite{yin2021center} regresses a confidence heatmap with the target defined by Gaussian kernels for anchor-free 3D detection.

To further improve the feature quality, LiDAR-image detectors fuse RGB images into point clouds to extract robust multi-modality features.
Early works like MV3D~\cite{chen2017multi} and F-PointNet~\cite{qi2018frustum} predict 2D region proposals to fuse image and point cloud features.
3D-CVF~\cite{yoo20203d} combines the camera and LiDAR features using the cross-view spatial feature fusion strategy.
PointPainting~\cite{vora2020pointpainting} projects LiDAR points into segmented images and appends the class scores to each point.
PointAugmenting~\cite{wang2021pointaugmenting} decorates point clouds with corresponding point-wise image features extracted by pre-trained 2D detectors.
S2M2-SSD~\cite{zheng2022boosting} transfers knowledge from LiDAR-image features and responses to a LiDAR-only detector for both high precision and efficiency.
Besides, a few works~\cite{nabati2021centerfusion,nobis2019deep,chadwick2019distant,kim2020low} attempt to fuse radar and image data, and many monocular/stereo works~\cite{liu2021autoshape,shi2021geometry,luo2021m3dssd,reading2021categorical,chen2020dsgn,ma2021delving} detect objects only on RGB images.

\begin{figure*}
\centering
\includegraphics[width=17.5cm]{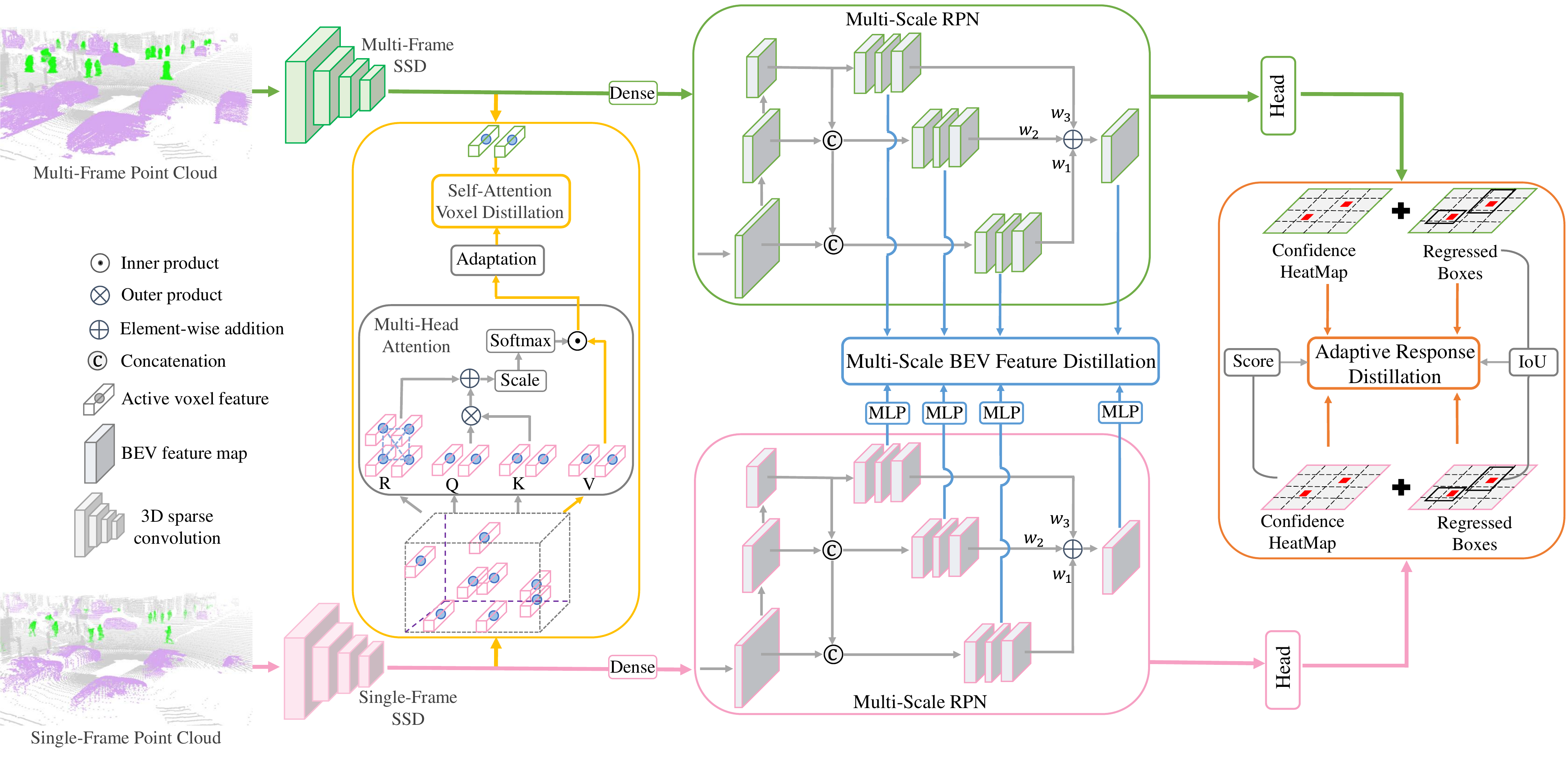}
\vspace*{-1.75mm}
\caption{
The pipeline of our SMF-SSD framework.
The single-frame SSD (bottom) takes a single-frame point cloud as input, whereas the multi-frame SSD (top) adopts a multi-frame one, in which we only densify ground-truth objects in the multi-frame point cloud by our designed multi-view dense object fusion (Section \ref{sec:3.2}).
To perform multi- to single-frame distillation, we first pre-train the multi-frame SSD (green arrows) and freeze its parameters.
We then train the single-frame SSD (pink arrows) to encourage it to learn to generate features and responses similar to those from the pre-trained multi-frame SSD.
Here, we design three levels of distillation techniques: (i) self-attention voxel distillation (Section \ref{sec:3.3}), (ii) multi-scale BEV feature distillation (Section \ref{sec:3.4}), and (iii) adaptive response distillation (Section \ref{sec:3.5}).
At testing (only pink arrows on the bottom), object detection can be done using the single-frame SSD and we only need a single-frame point cloud as input.
}
\label{pipeline}
\vspace*{-1mm}
\end{figure*}

To exploit multi-frame point clouds, 3D Auto Labeling~\cite{qi2021offboard} localizes objects in each frame and fuses the points of the predicted objects across multiple frames for further refinement.
3D-MAN~\cite{yang20213d} transfers multi-frame fusion from raw points to features for enhancing the single-frame features with multi-view features.
Besides, many works~\cite{li2019pu,yu2018pu,chen2019unpaired,xie2020grnet} propose various upsampling methods to attain high-density point clouds.
SPG~\cite{xu2021spg} generates semantic points at predicted foreground regions to recover missing object parts and merges semantic points with original points for detection.
PC-RGNN~\cite{zhang2020pc} designs an encoder-decoder architecture to densify the first-stage region proposals for refinement.

\textbf{Knowledge distillation.} \
Knowledge distillation~\cite{hinton2015distilling} was originally proposed to compress large models and later became widely adopted in image classification~\cite{yim2017gift,heo2019comprehensive,tung2019similarity}.
Recently, a few 2D detectors~\cite{chen2021distilling,guo2021distilling,dai2021general,qi2021multi,kang2021instance} decouple the BEV feature maps or focus on the significant instances for distillation.
In 3D object detection,~\cite{wang2020multi} first claims a multi- to single-frame distillation pipeline with a two-stage 3D detector but the technical details are not given.
Recently,~\cite{wang2021object} proposes a set-to-set distillation approach to transfer knowledge between object candidates.
Beyond these works, our SMF-SSD comprises a family of novel distillation techniques on intermediate voxels, BEV features, and responses to transfer knowledge from multi- to single-frame detectors, and sets a new state-of-the-art performance on single-frame 3D object detection.

\section{Simulated Multi-Frame SSD} 
\subsection{Overall Framework}
\label{sec:3.1}
Figure~\ref{pipeline} shows the pipeline of our SMF-SSD framework.
First, we pre-train a multi-frame SSD (top) on multi-frame point clouds with dense ground-truth objects.
We then freeze it
to produce features and responses for training the single-frame SSD (bottom) on single-frame point clouds.
The important goal of our framework is to train the single-frame SSD to learn to {\em generate simulated multi-frame features and responses\/} based on the outputs of the pre-trained multi-frame SSD.
To meet this goal, we design the multi-view dense object fusion and three levels of distillation techniques:
\begin{itemize}
\vspace*{-2.5mm}
\item
Multi-view dense object fusion (Section~\ref{sec:3.2}) densifies each single-frame ground-truth object by aggregating object points randomly sampled from multi-frame point clouds in the same sequence to form a multi-view dense object.
\vspace*{1.75mm}
\item
Self-attention voxel distillation (Section~\ref{sec:3.3}) builds an intermediate domain on single-frame voxels in ground-truth boxes with self attention to facilitate knowledge transfer from multi-frame voxels in a one-to-many manner.
\vspace*{1.75mm}
\item
Multi-scale BEV feature distillation (Section~\ref{sec:3.4}) extracts BEV features at different scales with our designed multi-scale RPN to distill low-level spatial features and high-level semantics separately to promote deep-feature consistency.
\item
Adaptive response distillation (Section~\ref{sec:3.5}) formulates adaptive weights in response distillation to activate single-frame responses of high confidence and accurate regressions.
\end{itemize}
Note also that both our single- and multi-frame SSDs adopt the open-source SOTA CenterPoint~\cite{yin2021center} as backbone.


\subsection{Multi-View Dense Object Fusion}
\label{sec:3.2}
To produce reliable features and responses, we enhance the input of the multi-frame SSD with multi-frame point clouds to reduce the sparsity and ambiguity of single-frame point clouds.
Yet, simply concatenating multi-frame point clouds in the same sequence leads to massive points, which may cause unaffordable computation overhead on the detector.
Hence, we propose to densify only ground-truth objects and avoid the dominated background, 
keeping the more significant foreground information for distillation.
To this end, we design the multi-view dense object fusion to generate dense ground-truth objects by multi-view composition
and produce random difference between the dense point sets of the same object appearing in sequential frames to promote sample diversity.

\begin{figure}
\centering
\includegraphics[width=8cm]{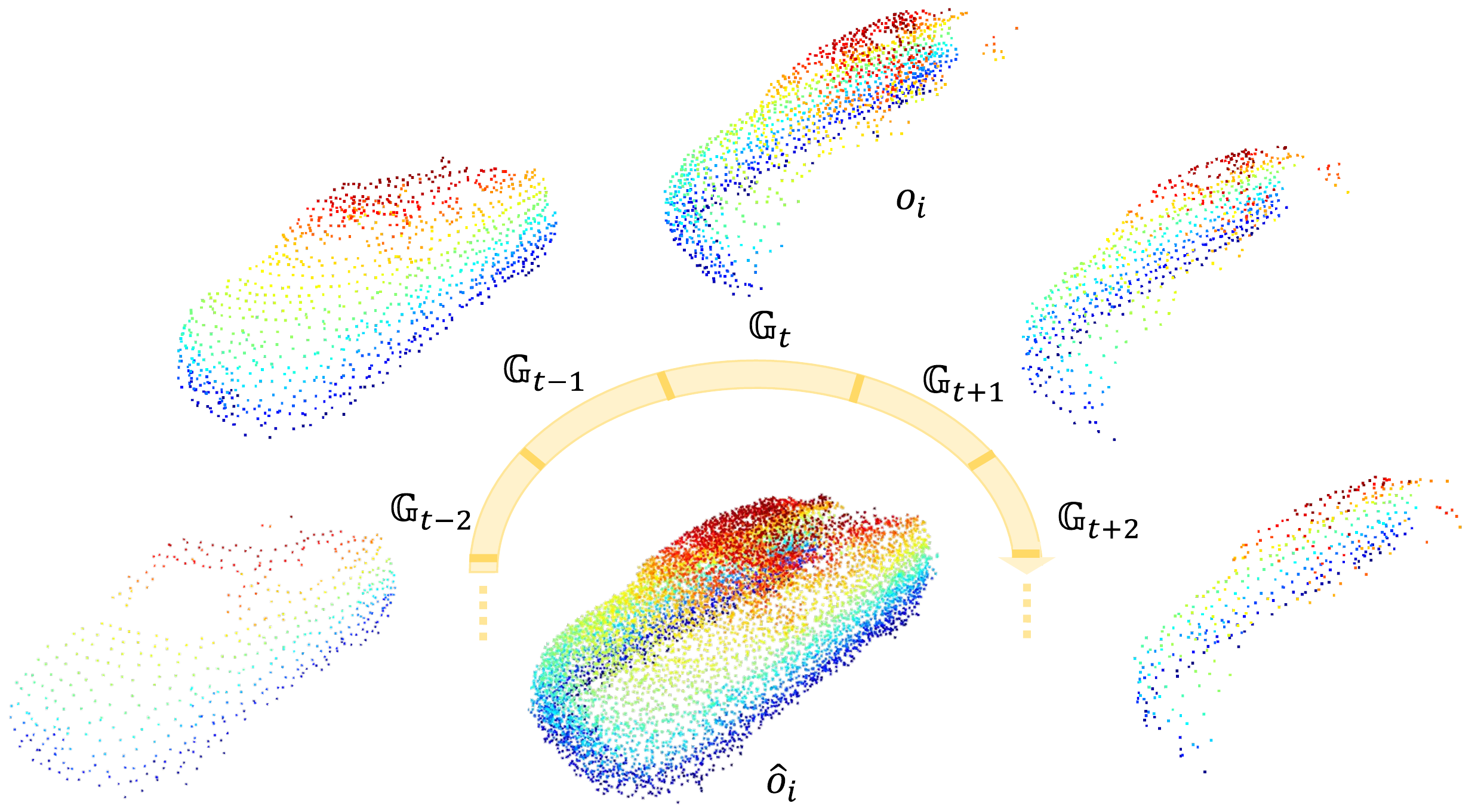}
\caption{
An example of multi-view dense object $\hat o_{i}$ produced from object points randomly sampled from multiple frame groups $\{\mathbb{G}_t\}_{t=1}^M$ using our multi-view dense object fusion.
}
\vspace*{-3mm}
\label{fig:dense}
\end{figure}

In detail, we denote $\{\mathbb{F}_{k}\}_{k=1}^N$ as the frames of all $N$ point clouds in a temporal sequence, where $N\!\!\approx$ 200.
Taking the $i$-th ground-truth object $o_i$ in the $j$-th frame $\mathbb{F}_j$ as an example, our fusion procedure has the following four steps.
First, from all the $N$ frames in $\{\mathbb{F}_{k}\}_{k=1}^N$, we form
$M$ local groups of frames $\{\mathbb{G}_t\}_{t=1}^M$, where $\mathbb{G}_t = \{ \mathbb{F}_{t'}, \mathbb{F}_{t'+1}, ..., \mathbb{F}_{t'+4} \}$; $M = \lfloor\frac{N}{5}\rfloor$; and $t' = 5t-4$.
For each group, say the $t$-th group, we calculate $N_{t_i}$, the average number of points per frame for object $o_i$, as object $o_i$ may appear in sequential frames.
Second, we concatenate all points of object $o_i$ in each group after a canonical transformation~\cite{shi2019pointrcnn} and sample $N_{t_i}$ points from it with furthest point sampling~\cite{qi2017pointnet++}, so the fused points of object $o_i$ in different frames can be randomized to be different point sets and
all object points in a group are fully exploited.
Third, we concatenate the object points sampled from all groups to generate a multi-view dense object $\hat o_{i}$ of $\sum_{t=1}^{M}\!N_{t_i}$ points, and then denoise it by removing around 0.5$\%$ points.
Last, we subsample $\hat o_{i}$ with grid sampling to reduce redundant points by voxelizing it with a grid size of [0.1, 0.1, 0.15]m along the x, y, and z axes, respectively, and keeping at most five points in each voxel.
Figure~\ref{fig:dense} shows a fusion example.

For training efficiency, we conduct the above four steps on all ground-truth objects in $\mathbb{F}_j$ separately and save all fused objects as a binary file $\mathbb{B}_j$.
So, we can load $\mathbb{F}_j$ and $\mathbb{B}_j$ to generate a multi-frame point cloud during the training, taking just slightly more time compared with loading only $\mathbb{F}_j$ for the single-frame SSD.

\subsection{Self-Attention Voxel Distillation}
\label{sec:3.3}
With the dense ground-truth objects, multi-frame voxels are able to provide more information than the single-frame ones for predicting object boundaries and semantic classes,
so it is significant to effectively transfer knowledge from multi- to single-frame voxels.
Existing methods~\cite{guo2021distilling,wang2019distilling,dai2021general,sun2020distilling} tend to directly impose consistency constraints between every pair of teacher and student features for one-to-one distillation.
This approach, however, ignores the significant domain gap between multi- and single-frame voxel features caused by the huge information gap between the dense multi-frame objects and the sparse single-frame objects, making it hard to train single-frame voxels to effectively simulate the multi-frame ones.
To alleviate this issue, we design self-attention voxel distillation to exploit the interrelation between single-frame voxels to build an intermediate domain,~\eg, exploiting different parts of same object to infer the missing parts or exploiting similar parts in different objects to enhance one another, aiming to reduce the domain gap with the multi-frame voxels and further enhance the knowledge transfer from multi- to single-frame voxels in a one-to-many manner.

In detail, we build an intermediate domain on single-frame voxels with self attention before comparing them with the multi-frame voxels.
Importantly, we focus on voxels in the ground-truth bounding boxes to comply with the fused dense objects and avoid the massive background voxels.
Procedure-wise, we first select ground-truth voxels ($V^s$=$\{v^{s}_{i}\}_{i=1}^{N^s}$) in the last sparse convolution layer of single-frame SSD (denoted by superscript $s$) for distillation, as they have rich semantics and keep the original 3D spatial information.
To match with $V^s$, we filter the last-layer voxels ($V^m$=$\{v^{m}_{i}\}_{i=1}^{N^s}$) of the same center coordinates as $V_s$ from the multi-frame SSD (denoted by superscript $m$).
Note also that we can always find a $v^{m}_{i}$ with the same coordinates as $v^{s}_{i}$, as the single-frame point cloud is a subset of the multi-frame one, and the same goes for voxels.
Next, we build structural dependencies on single-frame voxels $V^s$ with self attention to attain an intermediate domain for better simulating multi-frame voxel features. 
Meanwhile, the knowledge of a multi-frame voxel can be transferred to multiple interrelated single-frame voxels by distillation.
So, we adopt~\cite{vaswani2017attention} to transform each single-frame voxel $v^{s}_i$ to query $q_i$, key $k_i$, and value $v_i$ of $C$ channels linearly with $W^q$, $W^k$, and $W^v$, respectively:
\begin{equation}\label{eq1}
    \begin{split}
        & q_i = W^q\cdot v^{s}_i,\;\; k_i = W^k\cdot v^{s}_i,\;\; v_i = W^v\cdot v^{s}_i,\;\; r_{ij}=W^r\cdot d^{s}_{ij}, \\
    \end{split}
\end{equation}
where we additionally calculate the relative position $d^{s}_{ij}$ between $v^{s}_i$ and $v^{s}_j$, and encode it as a value $r_{ij}$ with $W^r$ to represent the spatial relation and geometric structure between $v^{s}_i$ and $v^{s}_j$.
Next, we multiply each pair of query $q_i$ and key $k_j$ to calculate their similarity as the attention weight $\alpha_{ij}$, in which we add $r_{ij}$ to consider the impact of spatial relation on the voxel-wise dependency:
\begin{equation}\label{eq2}
    \begin{split}
        & \alpha_{ij} = \frac{q_i\cdot k_j + r_{ij}}{\sqrt{C}}\; \rightarrow \; \hat\alpha_{ij} = \frac{e^{\alpha_{ij}}}{\sum_{j}e^{\alpha_{ij}}},  \\
    \end{split}
\end{equation}
where we scale the sum by dividing $\sqrt C$ and normalize $\alpha_{ij}$ with softmax to get the final attention weight $\hat \alpha_{ij}$.
Then, we weight each value $v_j$ with $\hat \alpha_{ij}$ and calculate the cumulative sum $\hat v^{s}_{i}$ to replace the original voxel features $v^{s}_{i}$: 
\begin{equation}\label{eq3}
    \begin{split}
        & \hat v^{s}_{i} = \sum_{j}^{N^s} \hat\alpha_{ij}\cdot v_j, \\
    \end{split}
\end{equation}
where the intermediate-domain voxel features $\hat v^{s}_{i}$ can extract complementary features from all single-frame voxels in $V^{s}$ to maximize the consistency with each multi-frame voxel.
Conversely, the semantic knowledge of each multi-frame voxel can be adaptively transferred to all single-frame voxels in a one-to-many manner.

Overall, we use eight attention heads as above and a feed-forward network (FFN) to encode the single-frame voxels with residual connections, in which we concatenate outputs of all attention heads and compress it to the same dimension as $V^{s}$ with a linear layer:
\begin{equation}\label{eq4}
        \widehat V^{s} = Attention^{8}(V^{s}, R^s) + V^{s}
\ \ \ \text{and} \ \ \ \widetilde V^{s} = FFN(\widehat V^{s}) + \widehat V^{s},
\end{equation}
where $\widehat V^{s}$=$\{\hat v^{s}_{i}\}_{i=1}^{N^s}$ and $\widetilde V^{s}$=$\{\tilde v^{s}_{i}\}_{i=1}^{N^s}$ denote the output of the self attention and FFN, respectively; and $R^s$ denotes $\{r_{ij}\}_{i,j=1}^{N^s}$.
Last, we calculate the mean square loss $\mathcal L_{vxl}$ between each pair of single-frame voxel $\tilde v^{s}_{i}$ and multi-frame voxel $v^{m}_{i}$ for distillation:
\begin{equation}\label{eq5}
    \begin{split}
        & \mathcal L_{vxl} = \frac{1}{N^{s}} \sum_{i=1}^{N^s} ||\tilde v^{s}_{i} - v^{m}_{i}||^{2}_{2}. \\
    \end{split}
\end{equation}

\begin{figure}
\centering
\includegraphics[width=7.7cm]{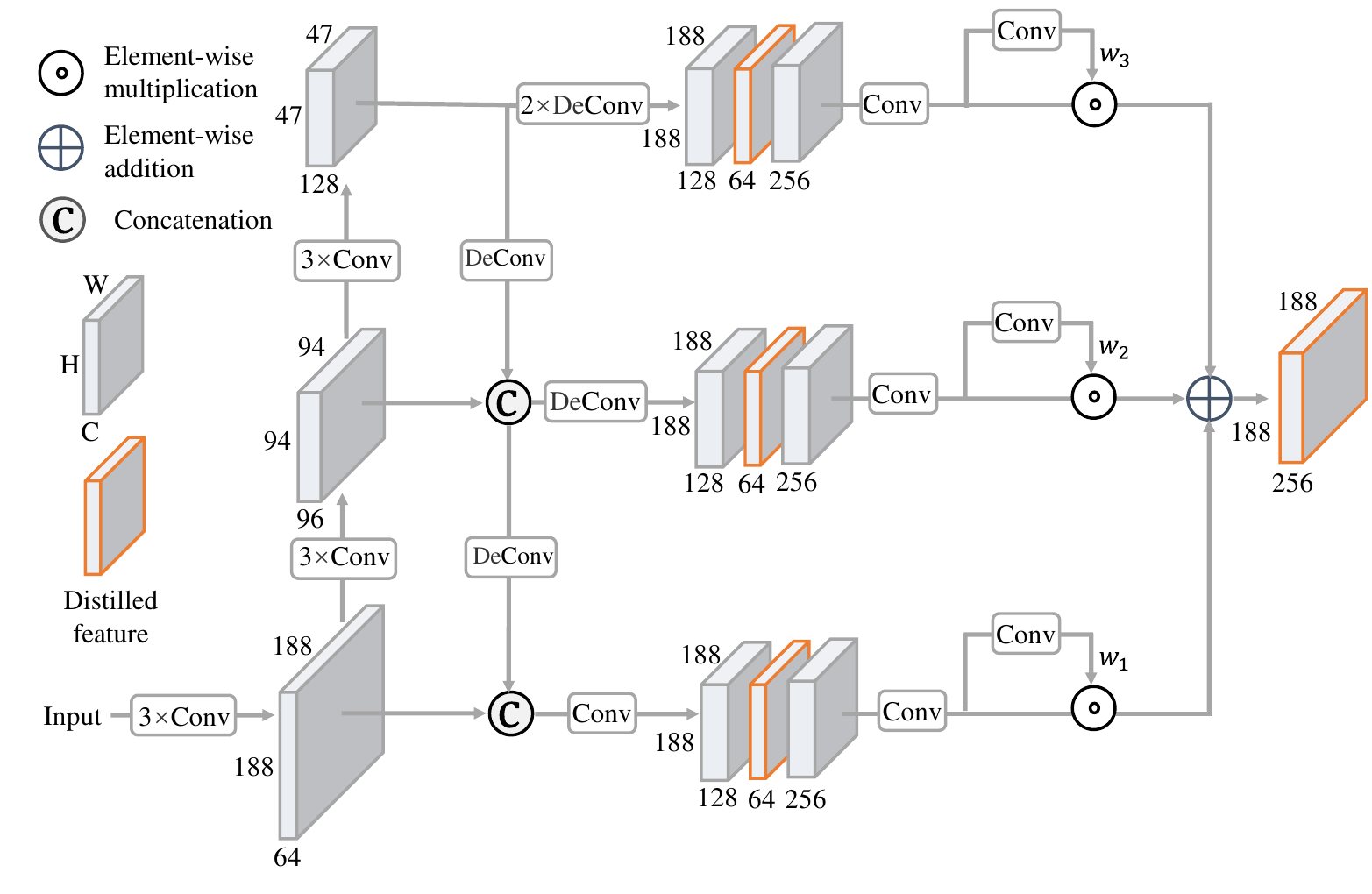}
\vspace*{-2.75mm}
\caption{
The structure of our designed Multi-Scale Region Proposal Network (MS-RPN),
which abstracts the input 3D features into three scales of BEV features and enriches low-level features with high-level semantics.
Before the final fusion, we design a bottleneck structure to reduce the feature channels for multi-scale BEV feature distillation.
}
\label{fig:MS-RPN}
\vspace*{-2mm}
\end{figure}


\subsection{Multi-Scale BEV Feature Distillation}
\label{sec:3.4}
With dense points as input, the multi-frame deep features can extract sufficient spatial information to describe the object boundaries and also rich semantics to recognize the object classes.
However, the original RPN in~\cite{yin2021center} composed of two convolution blocks cannot well present the spatial and semantic features separately and build dependencies between single-frame objects or object parts for distillation, due to limited feature scales and feature dependencies.
To address it, we design a new multi-scale region proposal network (MS-RPN) with more scales and cross-scale interactions,
and accordingly formulate a {\em multi-scale BEV feature distillation technique to separately transfer low-level spatial information and high-level semantics from multi- to single-frame BEV features\/}.

As shown in Figure~\ref{fig:MS-RPN}, our MS-RPN extracts BEV features in multiple scales to emphasize low-level spatial information or high-level semantics to different degrees.
Due to the cross-scale interactions and large receptive fields, it can also build dependencies between more single-frame objects or object parts to better simulate multi-frame BEV features.
In detail, we exploit three groups of Conv layers to produce scale difference and large stacked receptive fields, and further utilize the DeConv layers to enrich low-level features with high-level semantics for cross-scale interaction.
Next, all enriched features are upsampled to the input feature shape, followed by a bottleneck block to reduce the channels.
Last, we fuse all levels of features with spatial weights to attain the output.
Please find more implementation details in Section~\ref{sec:4.1}.
Note that our MS-RPN can only work by incorporating the multi-scale BEV feature distillation, so it cannot improve the single detector on its own.

With the bottleneck features extracted at different scales (see the orange-frame blocks in Figure~\ref{fig:MS-RPN}), we formulate the multi-scale BEV feature distillation to learn the differentiated knowledge in low-level spatial and high-level semantic BEV features:
\begin{equation}\label{eq6}
    \begin{split}
        & \mathcal L_{bev} = \sum_{l=1}^{4} \frac{1}{|B_{l}|} \sum_{i,j}^{H, W} \mathbb I(f_{ij} \in B_l) \cdot  ||\tilde f^{s}_{ij} - f^{m}_{ij}||^{2}_{2} \\
        \text{and} \ \
        & \tilde f^{s}_{ij} = MLP(f^{s}_{ij}), \\
    \end{split}
\end{equation}
where we transform the single-frame feature $f^{s}_{ij}$ to $\tilde f^{s}_{ij}$ with a multi-layer perception (MLP) and calculate the consistency loss between $\tilde f^{s}_{ij}$ and the multi-frame feature $f^{m}_{ij}$; and
$\mathbb I$ is an indicator function to filter $f^{s}_{ij}$ in the ground-truth bounding boxes from the set of level-$l$ features $B_l$, aiming to focus the distillation on the foreground features.
Besides, the bottleneck structure can reduce the feature channels significantly to compress the feature semantics into fewer channels, avoiding distillation on excessive channels and focusing on more significant channels for efficient knowledge transfer.


\subsection{Adaptive Response Distillation}
\label{sec:3.5}
Unlike the intermediate features, classification and regression responses of a detector directly impact the overall detection performance.
Given a denser amount of points, multi-frame responses usually have much higher precision than the single-frame ones, so we propose to take them as the soft targets to supervise the single-frame SSD.
In our framework, we design the adaptive response distillation to improve and activate the single-frame responses with high classification confidence and accurate regressed boxes.

To avoid massive background responses, we filter the ground-truth confidence heatmap $h^g$~\cite{yin2021center} with threshold $\tau$ to obtain the response positions close to the ground-truth object centers,~\ie, $G=h^g > \tau$.
With the estimated foreground positions $G$, we further calculate the adaptive weights to distill the classification responses:
\begin{equation}\label{eq7}
    \begin{split}
        & \mathcal L^c_{rsp} = \frac{1}{|G|}\sum_{i\in G} w^{c}_{i} \cdot \mathcal{L}_{sml1}|h^{s}_{i}-h^{m}_{i}| \\
        \text{and} \ \
        & w^{c}_{i} = \frac{h^{s}_{i} \cdot \sum_{j\in G} \mathcal{L}_{sml1}|h^{s}_{j}-h^{m}_{j}|}{\sum_{j\in G} h^{s}_{j} \cdot \mathcal{L}_{sml1}|h^{s}_{j}-h^{m}_{j}|},  \
    \end{split}
\end{equation}
where we multiply the adaptive weight $w^{c}_{i}$ with the smooth-$L1$ loss ($\mathcal{L}_{sml1}$) between the single- ($h^{s}_{i}$) and multi-frame ($h^{m}_{i}$) predicted confidence.
With the normalization, our adaptive weights keep the cumulative sum of original $\mathcal{L}_{sml1}$ loss unchanged, while highlighting the high-confidence responses predicted by the single-frame SSD.
Therefore, we can enhance the optimization of relatively accurate predictions to motivate the single-frame SSD to produce responses of high accuracy with higher priorities.
Similarly, we formulate the adaptive distillation on the regression responses as
\begin{equation}\label{eq8}
    \begin{split}
        & \mathcal L^r_{rsp} = \frac{1}{|G|}\sum_{i\in G} w^{r}_{i} \cdot \mathcal{L}_{sml1}|b^{s}_{i}-b^{m}_{i}| \\
        \text{and} \ \
        & w^{r}_{i} = \frac{IoU(b^{s}_{i}, b^{m}_{i}) \cdot \sum_{j\in G} \mathcal{L}_{sml1}|b^{s}_{j}-b^{m}_{j}|}{\sum_{j\in G} IoU(b^{s}_{j}, b^{m}_{j}) \cdot \mathcal{L}_{sml1}|b^{s}_{j}-b^{m}_{j}|},  \
    \end{split}
\end{equation}
where $\mathcal{L}_{sml1}$ calculates the average loss of all dimensions between each pair of bounding boxes predicted by the single- ($b^{s}_{i}$) and multi-frame ($b^{m}_{i}$) SSDs, and the adaptive weight $w^{r}_{i}$ is formulated by the $\mathcal{L}_{sml1}$ and $IoU$ between $b^{s}_{j}$ and $b^{m}_{j}$.
By this means, we can activate more accurate single-frame regressions by enhancing the optimization of the single-frame predicted boxes that are highly overlapped with the multi-frame ones.
Combining Eqs.~\eqref{eq7} and~\eqref{eq8}, we formulate the overall loss of our adaptive response distillation:
\begin{equation}\label{eq9}
    \begin{split}
        & \mathcal L_{rsp} = \pi_1 \cdot \mathcal L^c_{rsp} + \pi_2 \cdot \mathcal L^r_{rsp},\\
    \end{split}
\end{equation}
where we empirically set a larger $\pi_1$ than $\pi_2$ to focus more on the distillation of the classification responses.

\begin{table*}[t]
   \centering
   \footnotesize
   \caption{
   Comparison with the state-of-the-art single-frame detectors on the Waymo $test$ set.
   In both difficulty Level 1 and 2, our SMF-SSD attains the highest mAP(H) on all three classes consistently and new SOTA average mAP(H) of all three classes (denoted as "ALL").
   Compared to the single-frame SSD, our SMF-SSD also improves the mAP(H) significantly on all classes.
   In the table, `$\dagger$' means our reproduced single-frame results, whereas `*' means the SSDs built on our re-implemented CenterPoint.}
   \vspace*{-2mm}
   \resizebox{2.1\columnwidth}{!}{
   \begin{tabular}{c|c|c@{\hspace*{1.5mm}}c@{\hspace*{1mm}}c@{\hspace*{1mm}}c|c@{\hspace*{1.5mm}}c@{\hspace*{1mm}}c@{\hspace*{1mm}}c|c@{\hspace*{1.5mm}}c@{\hspace*{1mm}}c@{\hspace*{1mm}}c|c@{\hspace*{1.5mm}}c@{\hspace*{1mm}}c@{\hspace*{1mm}}c}
       \hline
       \multicolumn{1}{c|}{ \multirow{3}{*}{Method}} &
       \multicolumn{1}{c|}{ \multirow{3}{*}{Reference}} &

       \multicolumn{4}{c|}{VEHICLE} &
       \multicolumn{4}{c|}{PEDESTRIAN} &
       \multicolumn{4}{c|}{CYCLIST} &
       \multicolumn{4}{c}{ALL}
        \\ \cline{3-18}

       \multicolumn{1}{c|}{} &
       \multicolumn{1}{c|}{} &
       \multicolumn{2}{c}{Level 1} & \multicolumn{2}{|c|}{Level 2} &
       \multicolumn{2}{c}{Level 1} & \multicolumn{2}{|c|}{Level 2} &
       \multicolumn{2}{c}{Level 1} & \multicolumn{2}{|c|}{Level 2} &
       \multicolumn{2}{c}{Level 1} & \multicolumn{2}{|c}{Level 2} \\

       \multicolumn{1}{c|}{} &
       \multicolumn{1}{c|}{} &
       \multicolumn{1}{c}{mAP} & \multicolumn{1}{c}{mAPH} & \multicolumn{1}{|c}{mAP} & \multicolumn{1}{c|}{mAPH} &
       \multicolumn{1}{c}{mAP} & \multicolumn{1}{c}{mAPH} & \multicolumn{1}{|c}{mAP} & \multicolumn{1}{c|}{mAPH} &
       \multicolumn{1}{c}{mAP} & \multicolumn{1}{c}{mAPH} & \multicolumn{1}{|c}{mAP} & \multicolumn{1}{c|}{mAPH} &
       \multicolumn{1}{c}{mAP} & \multicolumn{1}{c}{mAPH} & \multicolumn{1}{|c}{mAP} & \multicolumn{1}{c} {mAPH} \\
       \hline
       \hline
         StarNet~\cite{ngiam2019starnet} &arXiv 2019
                        &61.50 &61.00 &54.90 &54.50 &67.80 &59.90 &61.10 &54.00 &- &- &- &- &- &- &- &- \\
         PPBA~\cite{cheng2020improving} &ECCV 2020
                        &67.50 &67.00 &59.60 &59.10 &69.70 &61.70 &63.00 &55.80 &- &- &- &- &- &- &- &- \\
         PointPillars~\cite{lang2019pointpillars} &CVPR 2019
                        &68.60 &68.10 &60.50 &60.10 &68.00 &55.50 &61.40 &50.10 &- &- &- &- &- &- &- &-   \\
         SA-SSD~\cite{he2020structure} &CVPR 2020
                        &70.24 &69.54  &- &- &57.14 &48.82 &- &- &- &- &- &- \\
         RCD~\cite{bewley2021range} &CoRL 2021
                        &71.97 &71.59 &65.06 &64.70 &- &- &- &-  &- &- &- &- &- &- &- &- \\
         Pesudo-Labeling~\cite{caine2021pseudo} &arXiv 2021
                        &74.00 &73.60 &- &- &69.80 &57.90 &- &- &- &- &- &-  &- &- &- &- \\
         M3DeTR~\cite{guan2022m3detr} &WACV 2022
                        &77.75 &77.17 &70.63 &70.06 &68.10 &58.87 &60.57 &52.37 &67.28 &65.69 &65.31 &63.75 &71.05 &67.09 &65.50 &61.92 \\
         Light-FMFNet~\cite{murhij2021real} &arXiv 2021
                        &77.85 &77.30 &70.16 &69.65 &69.52 &59.78 &63.62 &54.61 &66.34 &64.69 &63.87 &62.28 &71.24 &67.26 &65.88 &62.18 \\
         RangeDet~\cite{fan2021rangedet} &ICCV 2021
                        &75.83 &75.38 &67.12 &66.73 &74.77 &71.08 &68.58 &65.11 &64.59 &63.08 &61.93 &60.49 &71.73 &69.85 &65.88 &64.11 \\
         HIK-LiDAR~\cite{xu2021centeratt} &arXiv 2021
                        &78.63 &78.14 &71.06 &70.60 &76.00 &69.90 &69.82 &64.11 &70.94 &69.70 &68.35 &67.15 &75.19 &72.58 &69.74 &67.29 \\
         $\dagger$CenterPoint~\cite{yin2021center}&CVPR 2021
                        &78.18 &77.71 &70.63 &70.40
                        &76.59 &72.56 &70.71 &67.24
                        &70.53 &70.31 &67.95 &67.97
                        &75.10 &73.53 &69.76 &68.54\\
         \hline
               Single-Frame SSD*
                                          &- &77.76 &77.21 &70.33 &69.81
                                             &76.65 &73.75 &70.88 &68.11
                                             &71.25 &70.29 &68.97 &67.95
                                             &75.22 &73.75 &70.06 &68.63\\
               \textbf{SMF-SSD* (ours)}
                                         &-  &\bf79.57	&\bf79.10 &\bf71.98 &\bf71.55
                                             &\bf79.21	&\bf76.24 &\bf73.27 &\bf70.45
                                             &\bf72.42	&\bf71.35 &\bf70.04 &\bf69.00
                                             &\bf77.07	&\bf75.56 &\bf71.77 &\bf70.34 \\
         \rowcolor{LightCyan}
         \textit{Improvement}            & - &\textit{+1.81} &\textit{+1.89} &\textit{+1.65} &\textit{+1.74}
                                             &\textit{+2.56} &\textit{+2.49} &\textit{+2.39} &\textit{+2.34}
                                             &\textit{+1.17} &\textit{+1.06} &\textit{+1.07} &\textit{+1.05}
                                             &\textit{+1.85} &\textit{+1.81} &\textit{+1.71} &\textit{+1.71}  \\
      \hline
   \end{tabular}
   }
   \label{table1}
\end{table*}


\subsection{Overall Loss Function}\label{sec:3.6}
We train the single-frame SSD end-to-end by fixing the pre-trained multi-frame SSD (see Figure~\ref{pipeline}) and formulating the following supervision losses and distillation losses:
\begin{equation}\label{eq10}
  \mathcal{L}= \mathcal L_{cls} + \alpha \mathcal L_{reg} + \beta \mathcal L_{vxl} +  \lambda \mathcal L_{bev} +  \mu\mathcal L_{rsp}, \
\end{equation}
where $L_{cls}$ and $L_{reg}$ are classification and regression losses, respectively, as in~\cite{yin2021center}; and weights $\alpha$, $\beta$, $\lambda$, and $\mu$ are hyper-parameters.

\section{Experiments}
\subsection{Experimental Setup}\label{sec:4.1}
\textbf{Dataset.} \
We adopt the large-scale dataset Waymo~\cite{sun2020scalability} in our experiments due to its rich annotations on sequential frames of point clouds.
The dataset comprises 798/202/150 sequences that contain 158,081/39,987/29,647 sample frames for training/validation/testing.
Each sequence contains around 200 frames of point clouds, which are sampled with a 64-beam LiDAR sensor at 10Hz in 20 seconds.
All frames in every sequence are annotated with 3D object bounding boxes, so we can locate and densify ground-truth objects to produce the multi-frame point clouds.
For each frame, the point cloud has around 177K points on average and these points are distributed in full 360$^{\circ}$.
Each point in a frame contains the information of 3D coordinates and the intensity and elongation of laser pulse.
The 3D object detection task involves three classes of objects,~\ie, vehicle, pedestrian, and cyclist.
The goal is to localize each object with a bounding box of x, y, z, width, length, height, and yaw angle.

\textbf{Metrics.} \
We adopt the official metric of Waymo,~\ie, mean average precision (mAP) and mAP weighted by the heading accuracy (mAPH), to evaluate the detection performance.
The two metrics are calculated by matching the predictions with the ground-truth bounding boxes in two difficulty levels: Level 1 for boxes with more than five LiDAR points and Level 2 for boxes with at least one LiDAR point.
We compute the metrics of the two levels for each object class and also calculate the average metrics of all the classes in our experiments.
Besides, we break down each metric into multiple distance ranges to further analyze the detection performance.

\textbf{Network and Training.} \
We use a linear layer without bias for $W_q$, $W_k$, $W_v$, and $W_r$ in Equation~\eqref{eq1},
and two non-linear layers with ReLU for the FFN in Equation~\eqref{eq4}.
In Equation~\eqref{eq6}, the MLP consists of a convolution layer with a 1$\times$1 kernel and the ReLU function.
In Figure~\ref{fig:MS-RPN}, the strides of the DeConv layer and the first-layer of 3$\times$Conv are set as 2, and the kernels of all Conv and DeConv layers are set as 3$\times$3.
We compress all levels of features to single channel and normalize the concatenated three channels in each position with softmax to attain the spatial weights $w_1$, $w_2$, and $w_3$, which are multiplied with the corresponding features for the final addition.
To preprocess the input data, we set the detection range as [-75.2, 75.2]m for the x and y axes and [-2, 4]m for the z axis, and discretize the point clouds by a voxel size of [0.1, 0.1, 0.15]m along the x, y, and z axes, respectively.
The SSD models employed in our SMF-SSD framework are built on our implemented CenterPoint and MS-RPN, and we use the ADAM optimizer with the cosine annealing learning rate to train the model with a batch size of two.
We set $\tau$ as 0.1 in Section~\ref{sec:3.5}, $\pi_1$ and $\pi_2$ as 2.0 and 1.0 in Equation~\eqref{eq8}, and $\alpha$, $\beta$, $\lambda$, and $\mu$ as 2.0, 8.0, 1.0, and 1.0 in Equation~\eqref{eq9}.

\subsection{Comparison with the State-of-the-Arts}\label{sec:4.2}
We evaluated our SMF-SSD on the Waymo test set by submitting our predicted results to the official server.
Table~\ref{table1} shows the evaluated mAP and mAPH on all object classes in both level 1 and level 2 for comparison with state-of-the-art single-frame methods.
As shown in the table, our SMF-SSD attains the highest mAP and mAPH on all three classes in both level 1 and level 2 among all single-frame detectors, setting a new state-of-the-art performance in mAP and mAPH (see the "ALL" column) of all the three classes.
Compared with the baseline single-frame SSD, our SMF-SSD attains significant gains of +1.85, +1.81, +1.71, and +1.71 points on the average mAP and mAPH in both level 1 and level 2, respectively.
Also, our SMF-SSD demonstrates consistent improvements not only on vehicles but also on the pedestrian and cyclist classes.

\begin{table}[t]
\centering
\caption{Ablation study on our proposed modules: ``mf'' means our multi-view dense object fusion; ``voxel'', ``bev'', and ``rsp'' denote the distillation on intermediate voxels, multi-scale BEV features, and detection responses, respectively.
We report Level-2 mAPH of our SMF-SSD trained on 1/8 Waymo train set and evaluated on the complete val split.}
\vspace{-2mm}
\begin{tabular}{cccc|ccc|c}
    \hline
      \multicolumn{1}{c}{mf} &\multicolumn{1}{c}{voxel} &\multicolumn{1}{c}{bev} &\multicolumn{1}{c|}{rsp}  &\multicolumn{1}{c}{ \multirow{1}{*}{Veh}} &\multicolumn{1}{c}{ \multirow{1}{*}{Ped}} &\multicolumn{1}{c|}{ \multirow{1}{*}{Cyc}} &\multicolumn{1}{c}{ \multirow{1}{*}{All}}\\
      \hline\hline
       -                 &  -          &   -        &   -         & 57.74	 & 51.46   & 62.14	& 57.11 \\
       \checkmark        & \checkmark  &   -        &   -         & 58.89    & 51.63   & 63.12  & 57.88 \\
       \checkmark        &  -          & \checkmark &   -         & 58.45    & 52.28   & 62.53  & 57.75 \\
       \checkmark        &  -          &   -        & \checkmark  & 58.78    & 52.07   & 62.08  & 57.64 \\
       \checkmark        & \checkmark  & \checkmark &   -         & 59.13    & 52.59   & 63.27  & 58.33 \\
       \checkmark        & \checkmark  & \checkmark & \checkmark  & \bf 59.71    & \bf 54.25   & \bf 63.40  & \bf 59.12 \\
       -                 & \checkmark  & \checkmark & \checkmark  & 58.51    & 52.16   & 62.18  & 57.62 \\

    \hline
\end{tabular}
\label{table2}
\end{table}

In detail, SMF-SSD gives the largest improvements, for both mAP and mAPH of level 1 and level 2, on pedestrians,~\ie, 2.56, 2.49, 2.39, and 2.34 points, compared with other classes.
We think that the large improvement on pedestrians is because pedestrian objects can benefit more from the dense point clouds, as they often have sparser LiDAR points due to smaller sizes when compared with the other classes.
Compared with vehicles, the cyclist shows less gains,~\ie, +1.17, +1.06, +1.07, and +1.05 points, even though it tends to have sparser points.
We believe that it may be caused by the sample imbalance, as vehicle objects dominate the object samples and thus hinder the knowledge transfer for cyclist objects.

Since SMF-SSD only needs single-frame point clouds as input, it has high efficiency at inference compared to the multi-frame SSD, which needs to process multi-frame point clouds.
Also, the fusion of multi-frame point clouds is often time-consuming and depends on complex calibration and synchronization; see Section~\ref{sec:4.4} for the details.
Overall, our SMF-SSD can greatly boost the detection precision by learning from the multi-frame SSD, while maintaining high efficiency as the single-frame SSD at inference.

\subsection{Ablation Study}\label{sec:4.3}
To conduct ablation experiments efficiently on the large-scale Waymo dataset, we follow~\cite{wang2021pointaugmenting,shi2021pv} to generate a small representative training set by uniformly sampling 1/8 frames from the complete training set, and evaluate all results on the complete val split.
In Tables~\ref{table2}-\ref{table6}, we show the mAPH of Level 2 on all object classes to validate the effect of each module in our SMF-SSD.
In Table~\ref{table7}, we further explore the distance impact on the performance of our SMF-SSD.

\begin{table}[t]
\centering
\caption{Ablation study on self-attention voxel distillation with Level-2 mAPH reported: ``all voxel'' and ``gt voxel'' mean distilling all voxels and the voxels in ground truths, respectively; ``context'' means more "gt voxel" by enlarging the box dimension by 0.8m; our "self attention" is the voxel encoder to replace the common MLP; and ``Number'' means average number of voxels for distillation per sample.}
\begin{center}
\vspace{-2mm}
\begin{tabular}{c|c|ccc|c}
  \hline
  Distilled voxels                     & Number    & Veh   & Ped   & Cyc   & All   \\
  \hline
  baseline                             &-       & 57.74 & 51.46   & 62.14	& 57.11  \\
  all voxel                            &37867       & 57.76 & 51.47   & 62.15	& 57.12  \\
  gt voxel                             &576       & 58.18 & 51.53   & 62.44	& 57.38  \\
  + context                            &1305       & 58.23 & 51.56   & 62.51	& 57.43  \\
  + self attention                     &1305       & \bf 58.89    &\bf  51.63   &\bf  63.12  &\bf  57.88 \\
  \hline
 \end{tabular}
\label{table3}
\end{center}
\end{table}

\begin{table}[t]
\centering
\caption{Ablation study on multi-scale BEV feature distillation with Level-2 mAPH: we explore different stages of BEV feature distillation in original RPN and MS-RPN. Superscripts ``1'' and ``2'' mean 128 and 64 feature channels, respectively; ``rm inter'' means removing the cross-scale interactions in MS-RPN.}
\begin{center}
\vspace{-2mm}
\resizebox{0.95\columnwidth}{!}{
\begin{tabular}{c|c|c|ccc|c}
  \hline
  Type & Parameters           &Stage        & Veh   & Ped   & Cyc   & All   \\
  \hline
  \multirow {2}{*}{Raw RPN}
  {} & \multirow {2}{*}{4.58M} & baseline         & 58.25	 & 50.86   & 62.04	& 57.05 \\
  {} & {}                    & late dist          & 58.30	 & 50.88   & 62.03	  & 57.07 \\ \cline{1-7}
  \multirow {4}{*}{MS-RPN}
  {} & \multirow {4}{*}{2.35M} & baseline         & 57.74	 & 51.46   & 62.14	& 57.11 \\
  {} & {}                    & early dist         & 57.84    & 51.53   & 62.21  & 57.19 \\
  {} & {}                    & late dist          & 58.03    & 51.87   & 62.31  & 57.40 \\
  {} & {}                    & middle dist$^1$    & 58.24    & 52.09   & 62.48  & 57.60  \\
  {} & {}                    & middle dist$^2$    & \bf 58.45    & \bf 52.28   & \bf 62.53  & \bf 57.75  \\
  {} & {}                    & + rm inter    & 58.07    & 51.92   & 62.37  & 57.45  \\
  \hline
 \end{tabular}
}
\label{table4}
\end{center}
\end{table}

\textbf{Effect of self-attention voxel distillation.} \
As the first and second rows in Table~\ref{table2} show, our self-attention voxel distillation improves the Level-2 mAPH by 1.15, 0.17, and 0.98 points on vehicles, pedestrians, and cyclists, respectively.
The resulting average gain of 0.77 points contribute the largest single-module improvement, indicating the necessity of enhancing the single-frame voxel features.
Also, it can be seen that the mAPH gain on vehicles is the largest one among all classes, as the large-size vehicle often have more associated voxels and thus benefit more from voxel distillation.

Table~\ref{table3} shows further ablation results on voxel selection and self attention, in which we use a two-layer MLP to replace the self attention in 2nd-4th rows.
Comparing the first and second rows, we can see that distilling all background and foreground voxels cannot improve the mAPH effectively, as the massive background voxels hinder the foreground knowledge transfer.
While comparing the second and third rows, we can see a minor average gain of 0.26 points due to distillation only on foreground voxels, in which the reduced amount of voxels also greatly alleviate the computation overhead.
From the fourth row, we further see an average gain of 0.05 points by enlarging the ground-truth bounding boxes by 0.8m, thereby involving contextual voxels to train the single-frame voxels to be more context-aware.
Last, self attention on foreground voxels attains an average gain of 0.45 points, showing that the intermediate domain promotes foreground knowledge transfer.

\begin{table}[t]
\centering
\caption{Ablation study on adaptive response distillation with Level-2 mAPH: we compare distillation on all samples and ground-truth (``gt'') samples of classification (``cls'') and regression (``loc'') responses w/ and w/o adaptive weights.}
\begin{center}
\vspace{-2mm}
\begin{tabular}{c|ccc|c}
  \hline
  Distilled responses                        & Veh   & Ped   & Cyc   & All   \\
  \hline
  baseline                                   & 57.74 & 51.46   & 62.14	& 57.11  \\
  all                                        & 57.75 & 51.44   & 62.15	& 57.11  \\
  gt cls                                     & 57.86 & 51.67   & 62.15	& 57.22  \\
  gt loc                                     & 57.78 & 51.46   & 62.13	& 57.12  \\
  \hline
  adaptive gt cls                            & 58.46 & 51.89   & 62.12	& 57.49  \\
  adaptive gt loc                            & 57.95 & 51.74   & 62.11	& 57.26  \\
  adaptive gt cls + loc                      & \bf58.78    &\bf52.07   &\bf62.08  &\bf57.64 \\
  \hline
 \end{tabular}
\label{table5}
\end{center}
\end{table}

\textbf{Effect of multi-scale BEV feature distillation.} \
Comparing the first and third rows in Table~\ref{table2} shows that our multi-scale BEV feature distillation improves the Level-2 mAPH by 0.71, 0.82, 0.39, and 0.64 points on vehicles, pedestrians, cyclists, and the average, respectively.
Comparing the second and fifth rows in Table~\ref{table2} shows that it further boosts the mAPH by 0.24, 0.96, 0.15, and 0.45 points on top of the self-attention voxel distillation, showing the complementarity of the two modules.
Also, the mAPH gain on pedestrians is the highest, as the reduced difference in object scales on 2D feature maps narrows the gain gap of the two distillation modules.

Table~\ref{table4} shows further ablation study on the multi-scale BEV feature distillation.
Comparing the first and third rows, we can see that our MS-RPN cannot improve the single detector on its own.
Comparing the second and fifth rows, we can see that distilling the last-layer BEV features (``late dist'') leads to an average gain of 0.33 points in our MS-RPN while almost no improvement in the original RPN, validating the effectiveness of decomposing the BEV features into multiple scales for separate distillations.
%
%
%
We also distill multi-scales BEV features before the DeConv (``early dist'') and at the middle layer of bottleneck (``middle dist''), showing that compressing the middle features into 64 channels gains more.
Comparing the last two rows shows that mAPH drops when removing the cross-scale interactions in MS-RPN.
Comparing the numbers in second column shows that MS-RPN has only $\sim$ 2.35M parameters, about half of the original RPN, due to fewer feature channels.

\begin{table}[t]
\centering
\caption{Ablation study on multi-view dense object fusion, in which we report the Level-2 mAPH of the multi-frame SSD trained on our fused multi-frame point clouds. }
\vspace{-2mm}
\begin{tabular}{c|cccc}
    \hline
      \multicolumn{1}{c|}{Model} &\multicolumn{1}{c}{ \multirow{1}{*}{Veh}} &\multicolumn{1}{c}{ \multirow{1}{*}{Ped}} &\multicolumn{1}{c}{ \multirow{1}{*}{Cyc}} &\multicolumn{1}{c}{ \multirow{1}{*}{All}}\\
      \hline
       Single-Frame SSD      & 57.74	 & 51.46   & 62.14	& 57.11 \\
       Multi-Frame SSD       & 80.26     & 72.70   & 73.11  & 75.36 \\
    \hline
\end{tabular}
\vspace*{-0.5mm}
\label{table6}
\end{table}

\begin{table}[t]
   \centering
   \footnotesize
   \caption{Distance impact on our distillation method.
   We report Level-2 average mAPH in different distance ranges.
   Our SMF-SSD achieves consistent gains in all ranges.}
   \vspace{-2mm}
   \resizebox{0.99\columnwidth}{!}{
   \begin{tabular}{c|c|c|c|c}
       \hline
       \multicolumn{1}{c|}{ \multirow{1}{*}{Method}} &

       \multicolumn{1}{c|}{[0m, 30m]} &
       \multicolumn{1}{c|}{[30m, 50m]} &
       \multicolumn{1}{c|}{[50m, +inf]} &
       \multicolumn{1}{c}{Full Range} \\
       \hline
         SF-SSD      &73.50 &54.50	&35.05  &57.11  \\
         SMF-SSD     &75.32 &56.51  &37.02  &59.12  \\
         Gain        &+1.78   &+2.01   &+1.97   &+1.98 \\
      \hline
   \end{tabular}
   }
   \vspace*{-0.5mm}
   \label{table7}
\end{table}

\textbf{Effect of adaptive response distillation.} \
Comparing the first and fourth rows in Table~\ref{table2}, we can see that this module improves the Level-2 mAPH by 1.04, 0.61, -0.06, and 0.53 points on vehicles, pedestrians, cyclists, and the average, respectively.
Also, comparing the fifth and sixth rows shows that it further increases mAPH by 0.58, 1.66, 0.13, and 0.79 points on top of the previous two modules, validating the importance of simulating detection responses.

Table~\ref{table5} shows ablation studies on adaptive response distillation.
Comparing the first two rows, we can see that distilling all response samples cannot improve the baseline precision.
Also, distilling the ground-truth responses of either classification or regression gives minor improvements, as shown in the third and fourth rows.
With our formulated adaptive weights, we can see decent improvement in the last three rows, showing the necessity of activating the responses of high confidence and accurate localization.

\textbf{Effect of multi-view dense object fusion.} \
We use a pre-trained single-frame SSD to replace the pre-trained multi-frame SSD as the teacher for distillation, with results shown in the last two rows in Table~\ref{table2}.
It can be seen that our pre-trained multi-frame SSD contributes 1.2 (vehicle), 2.09 (pedestrian), 1.22 (cyclist), and 1.5 (all) more points than the pre-trained single-frame SSD.
In Table~\ref{table6}, we can see that the mAPH of our multi-frame SSD outperforms the single-frame SSD significantly, showing that our multi-view dense object fusion is effective to produce a high-performance teacher.

\textbf{Distance impact on distillation.} \
Next, we compare the Level-2 mAPH gains of our SMF-SSD in different distance ranges.
From Table~\ref{table7}, we can see consistent mAPH gains of around 2 points in all different distance ranges.
Since distant multi-frame objects also have fewer points than the nearby ones after the dense fusion, the information gain of dense objects at different distances against the corresponding single-frame objects are very close.

\subsection{Runtime Analysis}\label{sec:4.4}
Table~\ref{table8} presents a runtime analysis of our SMF-SSD, showing its high efficiency.
As shown in the table, SMF-SSD only needs 66.23ms to detect objects on a single-frame point cloud, while the multi-frame SSD needs 96.15ms to process a multi-frame point cloud, which contains dense objects and consumes more computation.
Also, fusing a multi-frame point cloud offline with multi-view dense object fusion takes around 21.34 seconds on average, so
it is impractical to obtain multi-frame point clouds online for training or testing the multi-frame SSD, while our SMF-SSD offers a promising solution to exploit multi-frame data without slowing down the final inference speed.
Note that the experiment was done on an Intel Xeon Silver CPU and a TITAN Xp GPU with a batch size of one.

\begin{table}[t]
\centering
\caption{Runtime (in millisecond) of our SMF-SSD vs. the multi-frame SSD, showing its high inference efficiency. Note that the multi-frame data fusion is performed offline instead of during the inference.}
\vspace{-2mm}
\begin{tabular}{cc@{\hspace{3mm}}c@{\hspace{3mm}}c@{\hspace{3mm}}c@{\hspace{3mm}}c@{\hspace{3mm}}c@{\hspace{3mm}}c}
                \hline
                \multicolumn{1}{c|}{Time}
                & Data Fusion      & Inference     & Overall     \\
                \hline
                \multicolumn{1}{c|}{Multi-Frame SSD}
                & 2.13e4   & 96.15   & 2.14e4   \\
                \multicolumn{1}{c|}{Single-Frame SSD}
                & -       & 66.23   & 66.23   \\
                \hline
            \end{tabular}
\label{table8}
\end{table}

\section{Conclusion}
We presented a novel SMF-SSD framework, showing that we can train a single-frame SSD and encourage it to produce high-quality features and responses by learning from a pre-trained multi-frame SSD.
This approach requires multi-frame point clouds only during training, and once trained, our SMF-SSD can efficiently detect 3D objects at high speed with single-frame point clouds as its only input.
Method-wise, we first propose a multi-view dense object fusion method to densify ground-truth objects to form a multi-frame point cloud.
We then propose a family of distillation techniques to transfer knowledge in our fused multi-frame point clouds to the single-frame detector:
self-attention voxel distillation from multi- to single-frame voxels in a one-to-many manner;
multi-scale BEV feature distillation for low-level spatial information and high-level semantics in BEV features; and
adaptive response distillation to activate single-frame responses of high confidence and accurate localization.
Experimental results on the large-scale Waymo dataset show that our SMF-SSD sets a new SOTA performance on both mAP and mAPH consistently for all object classes in both difficulty levels 1 and 2, while having high inference efficiency.

\vspace*{2mm}
\noindent
{\bf Acknowledgments.} \
This research is supported by project MMT-p2-21 of the Shun Hing Institute of Advanced Engineering (SHIAE), The Chinese University of Hong Kong (CUHK).



\bibliographystyle{ACM-Reference-Format}
\bibliography{egbib}


\end{document}